\documentclass{article}

    \PassOptionsToPackage{numbers, compress}{natbib}

    \usepackage[final]{neurips_2021}

\usepackage[utf8]{inputenc} 
\usepackage[T1]{fontenc}    
\usepackage{hyperref}       
\usepackage{url}            
\usepackage{booktabs}       
\usepackage{amsfonts}       
\usepackage{nicefrac}       
\usepackage{microtype}      
\usepackage{xcolor}         
\usepackage{multirow}
\title{Combining Data-driven Supervision with Human-in-the-loop Feedback for Entity Resolution}

\author{%
  Wenpeng Yin, Shelby Heinecke, Jia Li\\
  \textbf{Nitish Shirish Keskar, Michael Jones, Shouzhong Shi}\\
  \textbf{Stanislav Georgiev, Kurt Milich, Joseph Esposito, Caiming Xiong}\\
  Salesforce\\
  \texttt{\{wyin,shelby.heinecke,jia.li\}@salesforce.com} \\
}

\begin{document}

\maketitle

\begin{abstract}
The distribution gap between training datasets and data encountered in production is well acknowledged \cite{Bengio20, Koh21, Wang18}. Training datasets are often constructed over a fixed period of time and by carefully curating the data to be labeled. Thus, training datasets may not contain all possible variations of data that could be encountered in real-world production environments. Tasked with building an entity resolution system - a model that identifies and consolidates data points that represent the same person - our first model exhibited a clear training-production performance gap. In this case study, we discuss our human-in-the-loop enabled, data-centric solution to closing the training-production performance divergence. We conclude with takeaways that apply to data-centric learning at large.
\end{abstract}

\section{Introduction}

The distribution discrepancy between training datasets and data encountered in production is widely known \cite{Bengio20, Koh21, Wang18}. A common workflow for collecting training datasets is to curate data over a fixed period of time and then label them using annotators. This limits the number and kinds of variations that are present in a training dataset. Models relying only on training datasets may lead to underwhelming performance in production when confronted with perturbed or out-of-distribution samples. This training-production performance gap has been addressed more generally from several closely related perspectives including out-of-distribution learning \cite{Wang21}, model robustness \cite{Goel21RG}, and data augmentation \cite{Goel21}. Confronted with the training-production performance gap in a business setting, we found that an application-specific approach was needed.

 We were tasked with building an entity resolution system - a model that identifies and consolidates data points that represent the same person \cite{Christophides20}. For example, the model intends to recognize that a \texttt{Jane Doe} with address \texttt{123 Main St, San Francisco, CA} is the same person as \texttt{J. V. Doe} living on \texttt{123 Main St.} in Zipcode \texttt{94158}. Circumstantially, more information such as phone number or email address might be available too. Our model would be deployed in production and used as an automated data pre-processing step to potentially benefit many downstream analytics and modelling tasks. \textit{For this initial development phase, our model intends to classify identities based on one field - an individual's name and leaves the incorporation of other contextual fields in future phases}. For training data, we were given a highly curated set of labeled name pairs consisting of examples such as \texttt{(``Joe'', ``Joseph'', match)} or \texttt{(``Joe'', ``Jane'', mismatch)}. For evaluation, we had a diverse set of \emph{unlabeled} name pairs that included new instances of real-world name variations.  Our first model exhibited the training-production performance gap, inspiring us to create a human-in-the-loop enabled, data-centric approach for this application. 

In this case study, we discuss the details of our approach, a mix of data augmentation and data-derived rules, and the key takeaways applicable to data-centric learning more generally.  



\section{Challenges of Entity Resolution}\label{sec: challenges}
Across documents and records, the name of an individual can have different surface forms including capitalization, spellings, accents, ordering and others. We summarize some of the complex variances below.
(a)\textit{ Missing spaces/hyphens}: ``\texttt{Mary Ellen}'', ``\texttt{MaryEllen}'', and ``\texttt{Mary-Ellen}'' may represent the same person; (b) \textit{Initials}: a name, such as ``\texttt{James Earl Smith}'', could also be represented with initials such as ``\texttt{JE Smith}'', ``\texttt{J.E. Smith}'', ``\texttt{JE. Smith}'' or ``\texttt{J.E Smith}''; (c) \textit{Out-of-order components}: such as ``\texttt{Diaz Carlos Alfonzo}'' vs. ``\texttt{Carlos Alfonzo Diaz}'';
(d) \textit{Truncated name}: such as ``\texttt{Livingston Charles}'' vs. ``\texttt{Living Charles}''; (e) \textit{Nicknames}: a name's nickname  is often a shorter string (e.g., ``\texttt{Mike}'' vs. ``\texttt{Michael}'') or can  even be totally different by metaphone (e.g., ``\texttt{William}'' vs. ``\texttt{Bill}''); (f) \textit{Multilingual}: the users from different language backgrounds are allowed to type their names in native languages so that our system is expected to match multilingual names. For example, ``\texttt{Sch\"{u}tze}'' (German) matches with ``\texttt{Schuetze}'' (English); (g) \textit{Junk words}: when the system requires a user to type in her or his name, the user sometimes just puts a random string as the name, such as ``\texttt{too}'', ``\texttt{API}'', ``\texttt{such}'', etc; (h) \textit{Gender mismatch}: we should clearly reject the match of two names if they are typically used in different genders, such as ``\texttt{Daniel}'' (M) vs. ``\texttt{Daniela}'' (F).

\section{Training Dataset}
To provide supervision for developing a system, we manually annotated some data in different languages, including English, Arabic, French, German, Russian,  Spanish and Portuguese. The data consisted of the most popular names within our consumer data.  Also included were the most popular names by country (both male and female), as well as most popular nicknames.  We then created multiple permutations of these names to annotate different matching patterns. The detailed statistics are presented in Table \ref{tab:my_label}.

\begin{table}[!h]
    \centering
        \caption{Statistics of clean labeled data in different languages.}
    \begin{tabular}{lrrrrrrr}
    \toprule
         & English & French & German & Arabic & Russian & Spanish & Portuguese\\ \midrule
         \#\texttt{match} & 3,037& 396& 175& 845&510 & 1,023&659 \\
         \#\texttt{mismatch} & 129& 396&501 & 82&34 &601 &60\\ \bottomrule
    \end{tabular}

    \label{tab:my_label}
\end{table}

More matched pairs were included than mismatched pairs. To generate more mismatched instances, we used rules, such as deletion, insertion, and n-gram similarity, to collect about 20,000 mismatched pairs automatically. This labeled data is randomly split into \textit{train} (70\%), \textit{dev} (20\%) and \textit{test} (10\%).

\section{Baseline Approach}\label{sec:baseline}
The baseline is a system already in production. It is based on name transformation given predefined patterns: (i) \textit{Longest common substring pattern}, e.g., (``\texttt{J\"{o}rn}'' vs. ``\texttt{Jorn}'') $\rightarrow$ (``\texttt{j\"{o}2}'', ``\texttt{jo2}''), where digit ``2'' indicates the length of the common substring; (ii) \textit{Position unigrams metaphone pattern}, e.g., (``\texttt{aab}'' vs. ``\texttt{abbc}'') $\rightarrow$ (``\texttt{a1}'' vs. ``\texttt{b1c3}'').
So, given the input name pair (``\texttt{aab}'', ``\texttt{abbc}''), the following transformed pairs will be created: \texttt{[(``aab'', ``abbc''),  (``a2'', ``2bc''), (``a1'', ``b1c3'')]}. Then, in the training phase, each transformed pair will be mapped to the frequencies of belonging to the \texttt{match} and \texttt{mismatch} classes. For a test name pair, apply the same transformation extraction, and then feed into model, use the feature with highest probability to decide whether the input is a \texttt{match} or not.

Our goal was to build a model that outperforms this baseline on the held-out test set and real consumer data.   Next, we describe our human-in-the-loop enabled process for a more advanced system.

\section{System Development with Human-in-the-loop Evaluation}
Our  system was developed with the help of multiple rounds of human evaluations and feedback. We summarize the four versions in the following subsections.

\subsection{V1: ``pure deep learning system''}
\paragraph{System.} For our first version of system, we treat this name matching problem the same way as popular sentence matching tasks \cite{DBLPYinSXZ16}, such as paraphrase identification \cite{DBLPDolanQB04}, textual entailment \cite{DBLPDaganGM05}, etc. To be specific, we make use of the pretrained BERT model \cite{DBLPDevlinCLT19} \footnote{Initialized by the ``distilbert-base-multilingual-cased'' model.} to take two names as the input and conduct binary classification (i.e., \texttt{match} vs. \texttt{mismatch}). The probability corresponding to the \texttt{match} class is used as the matching score of two input names. Table \ref{tab:resultsround1} lists the results of our system V1 and the baseline. Due to the clear improvement on overall recall, our system outperforms the baseline with a large margin (F1: 95.07 vs. 70.46).

\begin{table}[!h]
\setlength{\tabcolsep}{3pt}
    \centering
        \caption{Comparison our systems (V1 and V4) with the baseline.}
    \begin{tabular}{ccccccccccc}
    \toprule
    
    & \multicolumn{7}{c}{Individual}& \multicolumn{3}{c}{Overall}\\\cmidrule(lr){2-8}\cmidrule(lr){9-11}
         & {English} & {French} & {German} & {Arabic} & {Russian} & {Spanish} & {Portuguese} & baseline & V1 & V4\\\midrule
         
         R & 94.71& 99.87& 85.96& 95.91&93.30 & 90.56&92.90 &56.00 & 94.55 & \textbf{98.48}\\
         P & 95.44& 100.0&100.0 & 97.40&95.46 &94.76 &99.10 & 95.00& 95.60 & \textbf{99.12}\\
         
         F1 & 95.07&99.93 & 92.45& 96.65& 94.37& 92.61& 95.90 &70.46 & 95.07& \textbf{98.79}\\
         \bottomrule
    \end{tabular}

    \label{tab:resultsround1}
\end{table}

\paragraph{Human evaluation and feedback.} As a part of the delivery pipeline, the V1 system was delivered to our engineering partners. Using approximately 2 million unlabeled name pairs of real customer data, the team checked for disagreements between the baseline system (Section \ref{sec:baseline}) and our proposed model. Those instances were then broken up based on their 10th percentiles in the predicted matching score of V1, randomly sorted those grouped percentiles and manually reviewed and labeled 5\% of the instances for each percentile group, which accounted for ~8k in total.  Despite the superiority of our V1 model in the clean data, the system exhibited unexpected error patterns:
 (a) \textit{Single letter/period match}, such as (``\texttt{.}'' vs. ``\texttt{jeff}''), (``\texttt{A}'' vs. ``\texttt{Edward}''); (b) \textit{Initial matches}, such as (``\texttt{A.J.}'' vs. ``\texttt{Deby}''), (``\texttt{T.J.}'' vs. ``\texttt{John Paul}''); (c) \textit{Common nicknames did not match}, e.g., (``\texttt{Susan}'' vs. ``\texttt{Susanna}''), (``\texttt{John}'' vs. ``\texttt{Johnnie}''); (d) \textit{Junk words match}, such as (``\texttt{Accounting}'' vs. ``\texttt{Debby}''), (``\texttt{Financial}'' vs. ``\texttt{Fay}''); (e) \textit{Double name with double name}, such as (``\texttt{Juan Manuel}'' vs. ``\texttt{Jean Michel}'').

We noticed that the real data contains instances that are much more diverse, such as the challenges discussed in Section \ref{sec: challenges}, than that in the clean data. Many error patterns do not have corresponding labeled data that can provide supervision, e.g., nickname recognition. 

\subsection{V2: deep learning with data augmentation}\label{sec:v2}
\paragraph{System.}To deal with the issues of V1, we try two threads: first, we conduct data augmentation specific to some error patterns, such as ``double name with double name'', ``Single letter/period match'', and ``Initial matches''; second,  we build our lists to cover junk words and nicknames, respectively. Specifically, our augmented data has 10k labeled examples for each error pattern; our junk work list has size 1,570, covering some stop words and some words that are clearly not used as person names; the nickname list consists of 3,458 name pairs in total. The augmented data is used to train the BERT model; the junk word list and the nickname list are used to determine the predictions: \texttt{match} if two names are in the nickname list, \texttt{mismatch} if one name is a junk word and the other is not equal to it.

\paragraph{Human evaluation and feedback.} The evaluation team used the same data as in V1, and the same way to sample pairs for human evaluation. The following new issues were reported: (a) \textit{Inconsistency}. Overall, our V2 system performed better than V1 for those patterns, this means our deep learning model (i.e., BERT) indeed learned supervision to some extent, but  inconsistently so. This is mainly related to patterns that were supposed to be solved by data augmentation, such as ``Initial matches'', ``Single letter/period match'', etc.  As an example: ``\texttt{Maria}'' is found to match with its initial ``\texttt{M}'' but ``\texttt{Dale}'' does not match with ``\texttt{D.}''. And some false positive cases were found: ``\texttt{Service Order}'' matches with ``\texttt{S.}''; (b) \textit{Gender mismatches}. For example, in V2 we found ``\texttt{Daniel}'' matches with ``\texttt{Daniela}'' and ``\texttt{Danielle}'' (both the latter are feminine). This means the training data cannot provide enough signal about name genders; (c) \textit{Lacking the Metaphone information}. For example, the system can neither recognize that ``\texttt{Christian}'' matches with ``\texttt{Kristian}'', nor ``\texttt{Pavel}'' matches with ``\texttt{Pawel}''.

\subsection{V3}
\paragraph{System.} To deal with the issues found for V2, we adopt the following solutions. (i) For the inconsistency issue, we rely on rules and prioritize rule-based decision over data-base learning (take the data-based decision only if the rules cannot handle it); (ii) For the gender mismatch, we first collect (name, gender) mapping from the Social Security website;  \footnote{\url{https://www.ssa.gov/oact/babynames/limits.html}} and (iii) We detect the metaphone\footnote{By ``org.apache.commons.codec.language.Metaphone''.} of names and compute name similarity based on that. As a part of the human evaluation and feedback, the following error patterns were provided: (a) \textit{Exact names do not match}. Our V3 system predicts ``mismatch'' for the pairs (``\texttt{.}'', ``\texttt{.}'') and (``\texttt{will}'', ``\texttt{Will}'') because V3 will detect them as junk words. However, the production team prefers to claim them \texttt{match} although they are not regular names; (b) \textit{False positive from machine learning predictions}. The system V3 tends to predict \texttt{match} for pairs (``\texttt{Cristina}'' vs. ``\texttt{Christian}''), (``\texttt{Marisa}'' vs. ``\texttt{Maria}''), etc., which were unable to be handled by rules.

\subsection{V4: the final system}
Our final system, i.e., V4, is derived by utilizing some simple while effective tricks: we use rules to handle the ``Exact names do not match''; for ``False positive from machine learning predictions'', those pairs were unable to be handled by rules, therefore the matching score fully comes from the BERT classifier. In the past versions, we used a threshold 0.5 to decide if an input pair is ``match'' or not. Based on the error cases, we tighten the threshold and fix it to 0.7. After the four rounds of iteration, our system V4 gets further improvement over the V1 (98.79 vs. 95.07 by F1 in Table \ref{tab:resultsround1}).

\paragraph{Remaining debatable issues.}  (a) \textit{One name as the sub-name of the other}, such as (``\texttt{Jose M}'' vs. ``\texttt{M}'') and (``\texttt{Jose Maria}'' vs. ``\texttt{Maria}''). Some researchers think the first pair should be \texttt{mismatch} while the second one be \texttt{match}; (b) \textit{Gender first or sub-name first?} Considering the pair (``\texttt{Jose Maria}'' vs. ``\texttt{Maria}'') again, ``\texttt{Jose Maria}'' is often used as a Spanish male first name, in this case, the prediction should be \texttt{mismatch} since \texttt{Maria} is a common female name; however, if we want to emphasize more on the sub-name pattern, the prediction should be \texttt{match}.  




\section{Related Work}
Most prior works, such as \cite{DBLPoloRWRS20,DBLPlaBNKGWVPSRM21,DBLKaushikKLY20,DBLottsWGK20,DBLdgenTWK20,DBLP2280,DBL5921},  tried to collect adversarial training data by human annotators to address the error patterns found in human-in-the-loop. We tried too but finally gave up due to two reasons:  (i) collecting large-scale adversarial data is even more costly than enlarging the training data with data augmentation; (ii) it performed worse than rules for some error patterns in which a simple rule can get high precision, such as gender-aware predictions.
\section{Discussion}
In this case study, we shared our human-in-the-loop-enabled, data-centric approach for building a production quality entity resolution model. In contrast to some research that work on clean labeled data and claim state of the art performance, we are trying to handle a real-world problem which inspires us to rethink the following questions:

\textbf{What is data?} By ``data'', the community usually refer to labeled or unlabeled data. Here, we advocate that rules are also a sort of precise expression of data. More specifically, If the issue can be exactly expressed by concise rules, that means rules are actually the objective status of model training on data. Therefore, we think ``data'' should be extended as ``data+rules''.

\textbf{How much data is needed?} Labeled data is always limited. However, the  challenges in the real world are often beyond the imagination. Data-centric learning can play a bigger role once combined with the human-in-the-loop interactions.

\bibliographystyle{plain}
\bibliography{neurips_2021}
\end{document}